\documentclass[10pt,twocolumn,letterpaper]{article}
 \pdfoutput=1

\usepackage{cvpr}
\usepackage{times}
\usepackage{epsfig}
\usepackage{graphicx}
\usepackage{amsmath}
\usepackage{amssymb}

\usepackage{booktabs}

\usepackage{algorithmic}
\usepackage{algorithm}
\usepackage{array}
\usepackage{textcomp}
\usepackage{stfloats}
\usepackage{url}
\usepackage{verbatim}
\usepackage{cite}
\usepackage{microtype}

\usepackage{url}

\usepackage{multirow, multicol}
\usepackage{float}
\usepackage{graphicx}
\usepackage{multirow, multicol}
\usepackage{tabulary,booktabs}
\usepackage{tablefootnote}
\usepackage{amsfonts}
\usepackage{color,soul}
\usepackage{multirow}
\usepackage{colortbl}
\usepackage{color, xcolor}
\usepackage{bm}
\usepackage[subtle]{savetrees}
\usepackage{comment}

\newcommand{\shortpaper}[1]{}
\newcommand{\remove}[1]{}
\newcommand{\removed}[1]{} 
\newcommand{\checkandremove}[1]{}

\usepackage{array}
\newcolumntype{H}{>{\setbox0=\hbox\bgroup}c<{\egroup}@{}}
\newcommand{\xhat}{\widehat{x}}
\usepackage{setspace}
\usepackage{epstopdf}
\usepackage{empheq}
\usepackage{multirow}
\usepackage{url}
\usepackage[title]{appendix}
\usepackage{enumitem}

\usepackage[mathlines]{lineno}



\usepackage{endnotes}
\usepackage{color}
\usepackage{xparse}

\newcommand{\E}{\text{E}}

\usepackage[breaklinks=true,bookmarks=false]{hyperref}

\cvprfinalcopy 

\def\cvprPaperID{32} 

\ifcvprfinal\fi
\begin{document}

\title{KL Regularized Normalization Framework for Low Resource Tasks}

\author{Neeraj Kumar\\
IIT Delhi\\
{\tt\small neerajkr2k14@gmail.com}
\and
Ankur Narang\\
IEEE Senior Member\\
{\tt\small annarang2@gmail.com}
\and
Brejesh Lall\\
IIT Delhi\\
{\tt\small brejesh@ee.iitd.ac.in}
}

\maketitle

\begin{abstract}
Large pre-trained models, such as Bert, GPT, and Wav2Vec, have demonstrated great potential for learning representations that are transferable to a wide variety of downstream tasks . It is difficult to obtain a large quantity of supervised data due to the limited availability of resources and time. In light of this, a significant amount of research has been conducted in the area of adopting large pre-trained datasets for diverse downstream tasks via fine tuning, linear probing, or prompt tuning in low resource settings.

Normalization techniques are essential for accelerating training and improving the generalization of deep neural networks and have been successfully used in a wide variety of applications. A lot of normalization techniques have been proposed but the success of normalization in low resource downstream NLP and speech tasks is limited. One of the reasons is the inability to capture expressiveness by re-scaling parameters of normalization. We propose Kullback-Leibler(KL) Regularized normalization (KL-Norm) which make the normalized data well behaved and helps in better generalization as it reduces over-fitting, generalises well on out of domain distributions and removes irrelevant biases and features with negligible increase in model parameters and memory overheads. Detailed experimental evaluation on multiple low resource NLP and speech tasks, demonstrates the superior performance of KL-Norm as compared to other popular normalization and regularization techniques.
\end{abstract}

\section{Introduction}

Due to breakthroughs in optimization techniques, big datasets, and streamlined designs of deep neural architectures, deep learning (DL) has gained significant success in a variety of domains. Nevertheless, deep learning is renowned for requiring huge labelled datasets, which restricts the scalability of a deep model due to the expense of annotation. Early research in this subject utilised data augmentation and regularisation approaches to mitigate the overfitting issue caused by a lack of data, but only to a limited degree.

With the coming of large pre-trained models in NLP such as Bert\cite{devlin2019bert}, GPT\cite{Radford2019LanguageMA}, Wav2Vec\cite{Baevski2020wav2vec2A}, Whisper\cite{Niranjan2021EndtoEndWT}, etc, low resource tasks have become an active area of research and have lots of industry use-cases such as manufacturing, gaming, metaverse, etc. A lot of techniques such as fine tuning, linear probing \cite{Kim2018BilinearAN}, prompt tuning \cite{Zhou2022LearningTP} have been explored to improve the performance of these model in low resource settings. No work has been done in the direction of normalization to make it work better for low resource tasks. 

The normalization method is one of the fundamental contributions to the deep learning community. It is a method of adaptive reparametrization, motivated by the difficulty of training very deep neural networks\cite{Goodfellow-et-al-2016}. Batch Normalization~\cite{Ioffe2015BatchNA} is the first normalization technique proposed to prevent the training from getting stuck in the saturated regimes of non-linearities and improves the training speed. It gives regularization effects which leads to better generalization of deep neural networks. It promotes larger learning rate  and reduces the photometric distortions as batch normalized networks train faster and observe each training example fewer times. 
 
Beyond Batch Normalization, several normalization techniques have been proposed which work better on various applications such as stylization~\cite{Ulyanov2016InstanceNT,Huang2017ArbitraryST,Nam2018BatchInstanceNF} , recurrent neural networks~\cite{Ba2016LayerN,Cooijmans2017RecurrentBN}, object detection~\cite{Wu2018GroupN} and faster convergence~\cite{Salimans2016WeightNA}, image to image translation~\cite{park2019SPADE}, neural acoustic modeling~\cite{Kim2017DynamicLN} and vision related applications~\cite{Kou2020StochasticN,Luo2021SwitchableNF}. Such techniques have gained success in all the fields of artificial intelligence such as NLP, vision, speech and others.
 
In the traditional batch normalization operation, normalization is done by mean shifting of feature map along with making it the unit variance. To increase the expressivity of network , we use two learnable parameters namely scale $\gamma$ and bias $\beta$ as shown in Equation \eqref{norma}. 
  \begin{equation}\label{norma}
      z = \gamma \odot \frac{x -\mu}{\sqrt{\Sigma}} + \beta
 \end{equation}

Assuming the input data is Gaussian, after passing through the batch normalization, it becomes the unit gaussian and to increase the representative power, we are adding the learnable parameters to shift its mean and increase it variance according to the deep learning tasks.

Now if we expand the Equation \eqref{norma} and combine the scale and shift with mean and variance, we will get the general equation of normalization(Equation \eqref{norma112}).
  \begin{equation}\label{norma23}
      z = \gamma \odot \frac{x}{\sqrt{\Sigma}} + (\beta - \gamma \odot \frac{\mu}{\sqrt{\Sigma}})
 \end{equation}
 
   \begin{equation}\label{norma112}
  z = \alpha \odot x +\zeta
   \end{equation}

Normalization makes the data well behaved by making it unit gaussian such that it helps in faster optimization, larger learning rate and implicit regularization. Equation \eqref{norma112} is the general formulation of normalization where the $\alpha$ and $\zeta$ incorporates the mean and variance which results in the batch normalization as shown in equation \eqref{norma}. There are various ways to make the normalization framework well behaved and effective for deep learning training with different formulation of $\alpha$ and $\zeta$.

We are looking into the normalization framework that works well in low resource setting as it has  a lot of industry and research relevance. High resource supervised datasets are time and resource consuming and sometimes impossible to scale in industry set up. With the coming of large pre-trained models such as Bert, Wav2Vec, etc. it is possible to achieve good results in low resource setting.

Current normalization approaches such as batch normalization, layer normalization have not been effective in low resource set up to make data well behaved, increasing the expressive power of the network and better generalization as shown in experimental section. No work has been done in making the data well behaved through normalization in low resource setting.

In this regard, we propose \textbf{\textit{KL-Norm}} \textit{KL Regularized normalization} framework which imposes the prior on the normalization framework through the KL divergence loss to follow gaussian distribution. This has shown promising result in low resource tasks. KL Regularized  normalization helps to improve accuracy as KL regularization based method acts as an ensemble of data. It helps in better training generalization as it reduces overfitting by adding a regularization loss function in the training schedule. It generalizes well on out of domain datasets as compared to other normalization techniques. It filters relevant features and removes the superficial features or biases present in the dataset or pre-trained model. An overview of the proposed normalization mechanism is shown in Figure 2. We specifically make the following contributions:

\begin{itemize}
    \item  We propose a novel KL Regularized normalization framework (KL-Norm) which incorporates rescaling parameter computation by considering regularization loss (Section ~\ref{sec:mathvarnorm}).
 
    \item KL-Norm demonstrates better expressiveness due to regularization loss and generalizes well by reducing over-fitting. It incorporates uncertainty which promotes better out of domain data generalization.

    \item Detailed experimental analysis demonstrates superior accuracy and performance of KL-Norm as compared to other normalization techniques, on low resource downstream NLP tasks including: sentiment classification, characterizing semantic relationships, semantic textual similarity, textual entailment and paraphrase detection as well as downstream speech task such as keyword detection and emotion classification. 

\end{itemize}

\section{Related Work}

We have divided this section into two parts namely low resource NLP and normalization techniques.
\paragraph{Low resource NLP and Speech}

Earlier works in low resource include feature engineering which requires significant efforts while adapting to the new datasets\cite{tan_empirical_2008}. Other work is transferring knowledge across the domain to increase the data require for training\cite{zoph_transfer_2016}. Adversarial training\cite{goodfellow_generative_2014} is one of these approach that uses the knowledge of domain where plentiful data is present and does the out of domain adaptation on low resource data\cite{ganin_domain-adversarial_2016}. However, these approaches have not used a pre-trained generic language model, but perform pre-training for each task individually. Another set of low resource training involves using language model\cite{chaudhari_entropy-sgd:_2017,izmailov_averaging_2018}. \cite{hao_visualizing_2019} showed that a classifier that fine-tunes a pre-trained BERT model generally has wider optima on the training loss curves in comparison to models trained for the same task from scratch, indicating a more general classifier. In the speech domain, pre-trained architectures such as DeepeSpeech\cite{Hannun2014DeepSS} and Wav2Vec\cite{Baevski2020wav2vec2A} have been  explored for various classification task as intent\cite{yadav2021intent} , phoneme \cite{Baevski2020wav2vec2A}, speaker and language identification\cite{Fan2021ExploringW2}. Another approach of linear probing is used in large pre-trained model where the linear layer is added on top of pre-trained model with fine tuning the linear layer only. Prompt tuning has gained a lot of attention after the coming of GPT-3 model where the discrete or continuous prompts are used to predict the tasks by the large pre-trained model. However, none of them studied the impact of normalization based on KL regularization in low resource setting.

\paragraph{Normalization techniques}
Batch normalization(BN) is the first form for normalization techniques proposed in ~\cite{Ioffe2015BatchNA} to normalize the feature maps by computing the batch statistics which helps in training the deep neural networks faster. Normalization works better as the first order optimization algorithms such as SGD works better on isotropic landscape(\cite{Nesterov2004IntroductoryLO}).

Motivated by this , a lot of normalization techniques have been proposed to deal different scenarios.Layer Normalization(LN) (\cite{Ba2016LayerN}) and Recurrent batch normalization(\cite{Cooijmans2017RecurrentBN}) give better performance in recurrent deep learning models. Instance Normalization(IN)~\cite{Ulyanov2016InstanceNT} and Adaptive Instance Normalization~\cite{Huang2017ArbitraryST} helps in image stylization , Group Normalization~\cite{Wu2018GroupN} improves performance in object detection ,Weight Normalization~\cite{Salimans2016WeightNA} speeds up the convergence by reparametrization of weight vectors in a neural network that decouples the length of those weight vectors from their direction.Batch-Instance Normalization(BIN)~\cite{Nam2018BatchInstanceNF} controls the styles adaptively to the task and selectively to individual feature maps.
SPADE~\cite{park2019SPADE} makes this denormalization spatially sensitive. SPADE normalization boils down to "conditional batch normalization which varies on a per-pixel basis". Switchable Normalization~\cite{Luo2021SwitchableNF} are dynamically select BN,LN and IN in proportion and works better in vision tasks. Stochastic Normalization ~\cite{Kou2020StochasticN} gives regulaization effects and better on vision tasks.\cite{Jia2019InstanceLevelMN} provides a meta learning mechanism for instance-level normalization techniques.\cite{Kim2017DynamicLN} generates the rescaling parameters by different speakers and environments for
adaptive neural acoustic modeling via Layer Normalization.The proposed KL-Norm uses KL Regularized inference based affine parameters to show the expressive power on low resource NLP downstraemed tasks.

\section{Theoretical Foundation of KL Regularized Normalization}\label{sec:mathvarnorm}
\subsection{Preliminaries : Batch Normalization}

Batch normalization(BN)~\cite{Ioffe2015BatchNA} is first introduced for faster convergence and training stability. In traditional deep networks, too-high learning rate may result in the gradients that explode or vanish, as well as getting stuck in poor local minima. 
BN helps address such issues. By normalizing activations throughout the network, it prevents small changes to the parameters from amplifying into larger and suboptimal changes in activations in gradients; for instance, it prevents the training from getting stuck in the saturated regimes of nonlinearities. It leads to better generalization of network because it has the implicit regularization effect and sometimes the neural netowrk do not requires the explicit regularization techniques such as dropout~\cite{srivastava2014dropout}, mixout~\cite{lee2019mixout}, weight decay~\cite{krogh1992simple}, etc

Equation~\eqref{norma} is the batch normalized output(z) with input($x_1 \cdots x_n$) is used to calculate the mean($\mu$) and variance($\sigma^2$).  Use of scale($\gamma$) and bias($\beta$) in Equation~\eqref{norma} give flexibility to work with normalized input($\hat{x}$), if there is a need, thus increasing the representation power. 

 \begin{equation}\label{mean}
      \mu = \frac{1}{m}\sum_{i=1}^m x_i  \quad \sigma^2 =  \frac{1}{m}\sum_{i=1}^m (x_i-\mu)^2
 \end{equation}

  \begin{equation}\label{norma}
      z = \gamma \odot \frac{x -\mu}{\sqrt{\sigma^2+\epsilon}} + \beta
 \end{equation}

At the inference stage, mini-batch estimations
$\mu$ and $\sigma^2$ are not available, so BN tracks moving average of the statistics during training(Equation~\eqref{batcnmoving}) where  $\alpha$ is the coefficient of moving average, $\hat{\mu}$ and $\hat{\sigma}$ moving average versions of $\mu$ and $\sigma$ . These moving statistics $\hat{\mu}$ and $\hat{\sigma}$ at iteration t are used to normalize the feature map as given in Equation~\eqref{inferbn} during inference.

\begin{equation}\label{batcnmoving}
      \hat{\mu}^{(t)} = \alpha \mu^{(t)} + (1-\alpha) \hat{\mu}^{(t-1)} \quad \hat{\sigma^{2(t)}} = \alpha \sigma^{2(t)} + (1-\alpha) \hat{\sigma^{2(t-1)}}
 \end{equation}

   \begin{equation}\label{inferbn}
      \hat{x} = \frac{x -\hat{\mu}}{\sqrt{\hat{\sigma^2}+\epsilon}}
 \end{equation}

 In the batch normalization setting the normalized feature map($\hat{x}$) follows the isotropic gaussian distribution\cite{Goodfellow-et-al-2016} with zero mean and unit variance i.e. $\hat{x} \sim \mathcal{N}(0,I)$. The reparamterization of the normalized feature map through rescaling paramters i.e. mean and variance(Equation \ref{eq:varnm}) allows the output to represent the same family of functions of the input as the old parametrization, but the new parametrization has different learning dynamics as shown in Equation \eqref{norma}. This setting is not found useful for low resource setting and not able to capture the expressiveness through normalization and unable to perform well in out of domain generalization.

 \subsection{KL Regularized Batch Normalization}

 We denote normalized feature map, $z^{i}$ $i=1\ldots,K$. To make the data well behaved and capture more representative power through normalization, we impose a prior, $t^{i}$ which will follow gaussian distribution. Assume the Cross-Entropy (CE) loss with respect to the neural network parameters $\theta$ is denoted by ${\cal L}(\theta)$. To impose prior on normalized feature map, we pose the optimization as a constrained problem: 
\begin{equation}
\begin{aligned}
\min\limits_{\theta}\quad & {\cal L}(\theta) \\
s.t. \quad & \E[ d_{z}(z^{i}, t^{i})]  \leq \epsilon_i, \ i=1,\ldots,K 
\end{aligned}
\end{equation}

This is equivalent to 
\begin{equation}
\min\limits_{\theta}\quad {\cal L}(\theta)+\lambda \sum\limits_{i=1}^{K} d_{z}(x^{i}, t^{i})
\end{equation}

We apply gradient-based optimization to the following loss (${\cal L}(\theta)$ is the CE loss):
In particular, we choose $\E[d_{z}(z^{i}, t^{i})]$ as KL loss between the normalized feature map and the gaussian prior. This will bring the distribution of normalized output closer to gaussian prior and add regularization effect in the network. The $\lambda$ is the hyperparameter which is tuned according to the task.

  In the proposed normalization setting, the affine parameters i.e. $\gamma$ and $\beta$ can be seen as the rescaling parameters i.e. mean($\mu_{v}$) and standard deviation($\Sigma_{v}^{\frac{1}{2}}$) of normalization framework. The rescaling parameters (mean and variance) can be modeled with deep neural network architecture. We have used multi layer perceptron(MLP) to model mean and variance. With this, the KL Regularized normalized output $y$ will defined by the Equation \ref{eq:varnm}.

   \begin{equation}\label{eq:varnm}
       \mu_{v} = MLP(x) \quad \Sigma_{v}^{\frac{1}{2}} = MLP(x)
 \end{equation}

  \begin{equation}\label{eq:varnm}
      z =  \Sigma_{v}^{\frac{1}{2}}\odot \xhat + \mu_{v} 
 \end{equation}
  
  We have imposed the Gaussian prior on normalized output though KL divergence loss function which is given by Equation \ref{eq:kllossaa} where $\mu_0$ and $\mu_1$ are $K-$dimensional mean vectors, and $\Sigma_0$ and $\Sigma_1$ are diagonal co-variance matrices. 
 
  \begin{equation}
  \begin{split}
\label{eq:kllossaa}
   \text{KL}(\mathcal{N}(\mu_0, \Sigma_0)\|\mathcal{N}(\mu_1, \Sigma_1)) = \frac{1}{2}(\text{tr}(\Sigma_1^{-1}\Sigma_0)+\\ (\mu_1-\mu_0)^T\Sigma_1^{-1}(\mu_1-\mu_0)-K+\log(\frac{\det(\Sigma_1)}{\det(\Sigma_0)})) 
     \end{split}
  \end{equation}
  
 KL loss has several properties which are useful for better expressivity, generalization in low resource setting. We will discuss each of these properties in the next subsections. 

\paragraph{Regularization effect of Kullbach-Leibler Loss \cite{asperti2020balancing}}

 Thus KL loss  acts as a regularizer term means ‘keep the representations $z$ sufficiently diverse’. If we don’t include the regularizer, the model can learn to cheat and give each datapoint a representation in a different region of Euclidean space. With the Kl loss, a gaussian prior will be imposed on normalization framework and which results in the data points to follow the Gaussian distribution with similar representation close together.

\paragraph{Ensemble effect}
KL Regularized normalization incorporates uncertainty in two ways~\cite{Lakshminarayanan2017SimpleAS}.
It is doubly stochastic in the sense that both the underlying feature representation($z$) \textit{and} the labels ($y$) are regarded as random variables. Conversely, Deep neural networks only regard the labels as being random variables.KL Regularized normalization has the ability to model both mean and variance in the label predictions by explicitly modeling the representation distribution

In most deep neural network, the output layer
corresponds to a distribution in which variance is a function of mean (e.g., a
binary classifier predicting $p$ for a class occurrence must also predict the
variance $p(1-p)$).
The stochasticity in the representation induces an effective ensemble of decoder predictions \cite{Gawlikowski2021ASO,Ovadia2019CanYT}. This ensemble effect of KL Regularized normalozation helps the model to achieve higher accuracy and reduces over-fitting.~\cite{Lakshminarayanan2017SimpleAS}.

\paragraph{Out of domain generalization}
The second source of uncertainty is provided by the KL divergence between the conditional distribution over latent space given
the input and the latent space defined by the learned marginal $p_\phi(z)$; i.e.
$\operatorname{KL}[q(z|x)||p(z)]$.
Here, the marginal effectively learns a density model for the data, albeit in the lowerF-dimensional, 
lower-information latent-space rather than the original input space.
Density estimation, whether explicit~\cite{Devries2018LearningCF} or
implicit~\cite{Kliger2018NoveltyDW}, has been shown to be useful for out-of-distribution detection.

\begin{algorithm}
\caption{KL Regularized Normalization}
\label{alg:VI-NORM}
\begin{algorithmic}[1]
\REQUIRE 
\begin{tabular}[t]{@{}l}
mini-batch feature maps of each channel \\
x = $\{x_i\}_{i=1}^m$;\\
moving statistics $\hat{\mu}$,$\hat{\sigma^2}$;\\
moving statistics update rate $\alpha\in (0,1)$
\end{tabular}
\ENSURE  z = KL-NORM(x)

\STATE {\bf Training:} 
\begin{itemize}
 \item $\mu$ $\leftarrow$ $\frac{1}{m}\sum_{i=1}^m x_i$ , $\quad$ $\sigma^2$ $\leftarrow$  $\frac{1}{m}\sum_{i=1}^m (x_i-\mu)^2$ $\quad$ // mini batch mean and variance
\item $\hat{x}$ $\leftarrow$ $\frac{x-\mu}{\sqrt{\sigma^2+\epsilon}}$ $\quad$  // normalize with mini-batch statistics
\item $\mu_{v}$ $\leftarrow$ = MLP($x$), $\quad$ $\Sigma_{v}^{\frac{1}{2}}$ $\leftarrow$ = MLP($x$) $\quad$ // rescaling paramters i.e. mean and variance.
\item $z$ = $\Sigma_{v}^{\frac{1}{2}}$ $\odot$ $\hat{x}$ + $\mu_{v}$ $\quad$ $\quad$ $\quad$ $\quad$ $\quad$ $\quad$    // KL Regularized normalized output feature map
\item $\hat{\mu}$ $\leftarrow$ $\alpha \mu + (1-\alpha) \hat{\mu}$ $\quad$ $\hat{\sigma^2}$ $\leftarrow$ $\alpha \sigma^2 + (1-\alpha) \hat{\sigma^2}$ $\quad$ // update estimations of moving statistics
\end{itemize}

\STATE {\bf Loss:}

\begin{itemize}
    \item Loss = CE + $\beta$ KL 
\end{itemize}

\STATE {\bf Inference:}

\begin{itemize}
 \item $\mu_{v}$ $\leftarrow$ = MLP($x$), $\quad$ $\Sigma_{v}^{\frac{1}{2}}$ $\leftarrow$ = MLP($x$).  $\quad$ // rescaling paramters i.e. mean and variance.

\item $z$ $\leftarrow$
$\Sigma_{v}^{\frac{1}{2}}$ $\odot$ $\frac{x-\hat{\mu}}{\sqrt{\hat{\sigma^2}+\epsilon}}$ + $\mu_{v}$ $\quad$ $\quad$ $\quad$  // KL Regularized normalization output feature map
\end{itemize}

\end{algorithmic}
\end{algorithm}

\begin{figure*}[h]
\centering
   \includegraphics[width=0.8\linewidth]{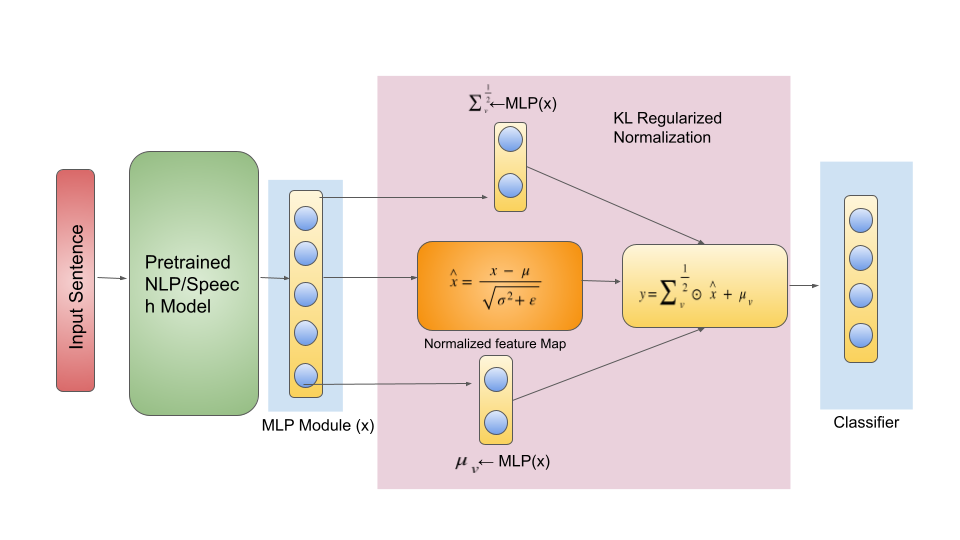}
   \caption{Architecture of proposed KL Regularized normalization framework }
   \label{fig:VINORMframe}
\end{figure*}

\paragraph{Algorithm}
Figure \ref{fig:VINORMframe} shows the proposed KL-Norm framework where the two multi-layer perceptron(MLP) layer is used to compute the rescaling parameters(mean($\mu_v$) and statistics($\Sigma_v^{\frac{1}{2}}$)). The normalized feature map($\hat{y}$) is calvulated using mean batch statistics to do affine transformation with rescaling parameters  .

We have proposed the algorithm of KL Regularized normalization(Algorithm \ref{alg:VI-NORM}). At the training stage , we compute the mean batch statistics($\mu$ and $\sigma^2$) to get the normalized feature map($\hat{x}$). The rescaling paramters i.e. mean($\mu_v$) and variance($\Sigma_v^{\frac{1}{2}}$) is calculated which goes into linear transformation to get the final output feature map($z$).The moving average statistics is calculated during training to be used at inference .The cross entropy(CE) and Kullbach-Leibler(KL) loss function is used while training with $\beta$ as a hyper-parameter. At the inference ,the moving average statistics calculates the normalized feature map. The final output is generated by linear transformation of normalized feature map using rescaling paramters.

\subsection{Model Architecture} 
Figure \ref{fig:VINORMframe} shows the architecture having pre-trained Language model, MLP module, KL Regularized normalization and classifier.The pre-trained BERT$_\text{Base}$ (12 layers, 110M parameters) and BERT$_\text{Large}$ (24 layers, 340M parameters) uncased ~\cite{devlin2019bert} implementation of~\cite{Wolf2019HuggingFacesTS} are used as our base models which gives 768 dimensional embeddings for sentence representation. The MLP module used to compute the compressed sentence representations  is a shallow MLP with $768$, $\frac{2304+K}{4}$, $\frac{768+K}{2}$ hidden units with a ReLU non-linearity, where $K$ is [$384,512$] gives K dimensional embedding. The MLP module acts as a bottleneck module for incorporating the relevant features in the KL Regularized normalization through the rescaling parameters.The KL Regularized normalization module include two linear layers which are used to calculate the affine/rescaling parameters to go into the normalization. A linear layer classifier is used on top to classify the sentence.Similar to~\cite{bowman2016generating}, we use a linear annealing schedule for $\beta$ and set it  as $\min(1,~\text{epoch}\times \beta_0)$ in each epoch, where $\beta_0$ is the initial value. 

For the speech related downstream tasks, we have used encoder of Wav2Vec2.0 pre-trained model as base model and added 1d convolution layer with 768 output channel. We have added MLP module having linear layers to compute the affine parameters of KL-Norm. A linear layer classifier is used on top to classify the speech related classes.

\subsection{Comparison methods}

We have used the common normalization techniques used in NLP tasks such as batch normalization(BN)\cite{Ioffe2015BatchNA}, layer normalization(LN)\cite{Ba2016LayerN} and group normalization(GN)\cite{Wu2018GroupN}. Such techniques do faster convergence and generalize well on NLP datsets. Apart from that, other common regularization techniques have also been used for the comparison namely  Dropout \cite{srivastava2014dropout} and Weight~\cite{krogh1992simple} Decay. While experimenting, we replace the KL Regularized normalization with the  comparison methods(Figure \ref{fig:VINORMframe}). We have also performed the experiments with BERT$_\text{Base}$ model given in Figure \ref{fig:VINORMframe} without KL Regularized normalization.

\begin{table*}[t] 

 \caption{Average results and standard deviation in parentheses over 5 runs on low-resource data in GLUE. $\bm{\Delta}$ shows the absolute difference between the results of the KL-Norm model with BERT-base.}

\centering
    \resizebox{12cm}{\height}{
    \begin{tabular}{lllllll} 
        \toprule 
         & \multicolumn{2}{c}{\textbf{MRPC}} & \multicolumn{2}{c}{\textbf{STS-B}} & \multicolumn{1}{c}{\textbf{RTE}}\\ 
         \cmidrule(r){2-3} \cmidrule(l){4-5}  \cmidrule(l){6-6}
        {\vspace{-0.75em} \textbf{Model}} &  \multicolumn{2}{c}{} & \multicolumn{2}{c}{} &\multicolumn{1}{c}{} \\ 
                 & \textbf{Accuracy} & \textbf{F1}  &\textbf{Pearson} & \textbf{Spearman} & \textbf{Accuracy}\\  
    \toprule 
    BERT$_\text{Base}$ & 84.31 (0.2) & 87.01 (0.2) & \textbf{84.43 (0.2)} & \textbf{83.28 (0.1)} & 64.23 (1.8) \\ 
    ~~+BN~\cite{Ioffe2015BatchNA} & 86.76 (0.5) & 90.49 (0.5) & 82.28 (0.5) & 81.57(0.7) & 66.42 (1.4) \\
    ~~+LN~\cite{Ba2016LayerN} & 85.23 (0.3) & 88.12 (0.7) & 84.33 (0.9) & 82.98(1.0) & 65.17 (0.9) \\
   ~~+GN~\cite{Wu2018GroupN} & 85.01 (0.2) & 87.98 (0.5) & 82.76 (0.8) & 81.91(1.1) & 65.55 (0.2) \\
    ~~+Dropout~\cite{srivastava2014dropout} & 85.55 (0.6) & 88.47 (0.2) & 84.11 (0.8) & 82.65 (0.7) & 65.12 (0.9) \\
    ~~+WD~\cite{lee2019mixout} & 85.01(0.2) & 86.91(0.2) & 84.02(0.8) & 82.29(0.5) & 65.02(0.8) \\
    \midrule
    KL-Norm$_\text{Base}$ & \textbf{87.25 (0.1)} & \textbf{91.03 (0.6)} & 82.49 (0.9) & 81.63 (0.8) & \textbf{70.42 (0.8)} \\
    $\bm{\Delta}$ & +2.94 & +4.02 & -1.94 & -1.65 & +6.19\\ 
    \toprule
     BERT$_\text{Large}$ &86.76 (0.7) & 90.20 (1.3) & \textbf{86.27 (0.3)} & \textbf{85.21 (0.1)} & 67.11 (0.8)\\ 
    ~~+BN~\cite{Ioffe2015BatchNA} & 85.53 (0.5) & 85.44 (0.5) & 84.47 (0.3) & 82.73(0.7) & 66.7 7(1.4) \\
        ~~+LN~\cite{Ba2016LayerN} & 86.27 (0.5) & 86.41 (0.5) & 86.52 (0.2) & 85.73(0.3) & 67.23 (1.4) \\
        ~~+GN~\cite{Wu2018GroupN} & 85.72 (0.3) & 85.87 (0.2) & 85.10 (0.3) & 85.21(0.7) & 66.2 (0.9) \\
      ~~+Dropout~\cite{srivastava2014dropout}&  86.51 (0.4) & 89.98 (0.2) & 86.13 (0.8) &85.02 (0.5) & 67.01 (0.6) \\ 
      ~~+WD~\cite{lee2019mixout} & 86.11(0.5) & 89.03(0.4) & 86.18(0.2) & 85.10(0.1) & 68.07(0.9) \\ 
     \midrule 
      KL-Norm$_\text{Large}$ & \textbf{87.99 (0.4)} & \textbf{91.38 (0.6)} & 86.01 (0.7) & 84.68 (0.9) & \textbf{71.01 (0.8)} \\
    $\bm{\Delta}$ & +1.23 &  +1.18 &  -0.26 &  -0.53 & +3.9 \\ 
    \bottomrule
    \end{tabular}}
    \vspace{-0.5em}
\label{tab:glue_results}
 
\end{table*}

\subsection{Training Details}

We have used the pre-trained BERT$_\text{Base}$ and BERT$_\text{Large}$ uncased models as the base models to see the effectiveness of the proposed method on down-streamed NLP tasks . We use the default hyper-parameters of BERT, i.e., we use a sequence length of $128$, with batch size $8$. We use the stable variant of the Adam optimizer~\cite{zhang2020revisiting,mosbach2020stability} with the default learning rate of $2\mathrm{e}{-5}$ through all experiments. We do not use warm-up or weight decay. 

We have used the Wav2Vec 2.0 \cite{Baevski2020wav2vec2A} pre-trained on LibriSpeech 960 hours as the. base model to see the. performance of proposed method on downstream speech. related tasks. We. have chosen Adam optimiser and set the learning rate of the backbone to $1\mathrm{e}{-5}$ for fine tuning and an L2 penalty of $1\mathrm{e}{-5}$ was established.

\subsection{Analysis of Increased Expressive Power}\label{sec:ecpress}

\paragraph{Glue Benchmarks}\label{sec:glue_results}
We have used the three low resource datasets namely MRPC and RTE for the evaluation of the proposed method. Table \ref{tab:glue_results} shows the comparison of the proposed model with the compared models having BERT-base uncased as the base model.  It shows that the proposed normalization model outperforms the other method in various evaluation metrics. Our KL-Norm method substantially improves the results and surpasses the prior work in all settings for both BERT$_\text{Base}$ and BERT$_\text{Large}$ models.  Due to the computational overhead of BERT$_\text{Large}$, for the rest of this work, we stick to BERT$_\text{Base}$.

\paragraph{Low Resource varying datasets} \label{sec:lowres}
We have used the four large NLP datasets such as  SNLI, MNLI, QNLI, and YELP  and subsample the dataset using the random seeds. We then evaluated the performance of NLI datasets under varying sizes of training data (200, 400, 600 800 and 1000 samples). We reported the average and standard deviation across 5 different seeds. Table \ref{tab:sampled_results_test} shows that KL-Norm consistently outperforms all the baselines on low-resource scenarios. 

We have also performed the experiments on three large speech related datasets such as Google Speech Command, Crema-D and ESD. We have evaluated the performance under varying size using random seeds. Table \ref{tab:sampled_results_test_speech}  have shown the that. KL-Norm performs better as compared to other methods.

\begin{table*}[t] 
    \caption{Test accuracies in the low-resource setting on text classification and NLI datasets under varying sizes of training data (200, 400, 600, 800 and 1000 samples). $\bm{\Delta}$ shows the absolute difference between the results of the KL-Norm model with BERT-base.} 
    \centering
    \resizebox{12cm}{\height}{
    \vspace{-1em}
    \begin{tabular}{l|l|l|l|l|l|l|l}
    \toprule 
\textbf{Data}     &  \textbf{Model} & $\bm{200}$ & $\bm{400}$& $\bm{600}$ & $\bm{800}$ & $\bm{1000}$ \\
    \toprule      
    \multirow{6}{*}{SNLI} & BERT$_\text{Base}$ & 60.07 (0.6)  & 66.87 (0.4)&68.69 (0.9) & 72.47 (0.8) & 72.98 (0.4) \\ 
    &~~+BN & 58.43 (0.4)  & 66.29 (0.5)&68.76 (0.4) & 71.96 (0.2) & 73.43 (0.5)  \\
    &~~+LN & 59.95 (0.3)  & 65.26 (0.4)&68.68 (0.6) & 72.52 (0.1) & 73.01 (0.2)  \\
    &~~+GN & 58.60 (0.2) & 66.11 (0.6)  & 67.57 (0.5) & 72.01 (0.3) & 73.10 (0.2)  \\
    &~~+Dropout & 58.44 (0.4)  & 66.76 (0.9)&68.74 (0.7) & 72.12 (0.5) & 72.58 (0.7)  \\
    &~~+WD & 59.23 (0.6) & 66.11 (0.9) & 68.41(0.8) & 72.48 (0.6) & \textbf{72.52} (0.3) \\ 
    \cmidrule(r){2-8}
    &~~+KL-Norm &\textbf{62.55} (1.3) &\textbf{69.02} (0.6) & \textbf{70.47} (0.3) & \textbf{73.92} (0.2) & \textbf{74.05} (0.7)\\
    & $\bm{\Delta}$ &  +2.48 &  +2.15 &  +1.78 &  +1.45 & +1.07& \\
    \toprule 
     \multirow{6}{*}{MNLI} & BERT$_\text{Base}$ & 45.53 (1.1) & 51.12 (0.9)  & 57.74 (0.8) & 58.83 (0.3) & 60.19 (0.7) \\
    &~~+BN & 46.34 (0.4)  & 52.14 (1.3)&55.7 (0.8) & 56.5 (0.5) & 59.5 (0.7)  \\
    &~~+LN & 45.25 (1.3) & 51.42 (1.3)  & 57.79 (0.4) & 59.03 (0.6) & 60.32 (0.7)  \\
    &~~+GN & 44.17 (1.8) & 50.53 (1.1)  & 57.34 (0.6) & 58.60 (0.3) & 59.17 (0.5)  \\
     &~~+Dropout & 45.44 (0.8) & 51.32 (1.1)  & 57.65 (1.1) & 59.43 (0.4) & 60.08 (0.9)   \\ 
    &~~+WD& 45.72 (0.7) & 51.42 (0.4)  & 57.34 (0.7) & 58.88 (0.6) & 60.24 (0.4) \\
     \cmidrule(r){2-8}
     &~~+KL-Norm & \textbf{47.65} (0.7)  & \textbf{53.20} (1.4) & \textbf{59.29} (1.2) & \textbf{60.17} (0.9) & \textbf{61.24} (0.8) \\
     & $\bm{\Delta}$ & +2.12 & +2.08 & +1.55 & +1.34 & +1.05\\ 
    \toprule 
    \multirow{6}{*}{QNLI}  &BERT$_\text{Base}$& 71.12 (0.6) & 75.30 (0.4)  &  75.9 (0.8) & 78.18 (0.2) & 79.51 (0.4)  \\
    &~~+BN & 71.70 (0.2) & 74.13 (0.5)  &  75.7 (0.3) & 77.12 (0.4) & 77.32 (0.4)  \\
    &~~+LN & 71.95 (0.4) & 74.97 (0.8)  &  75.8 (0.3) & 78.38 (0.2) & 79.67 (0.4)  \\
    &~~+GN & 71.24 (0.4) & 74.16 (0.6)  & 75.7 (0.2) & 77.55 (0.1) & 77.45 (0.3)  \\
    &~~+Dropout & 71.43 (1.2) & 74.77 (0.4)  &  75.23 (0.7) & 78.67 (0.5) & 79.23 (0.3)  \\
    &~~+WD & 71.43 (0.3) & 74.61 (0.6)  &  75.51 (0.2) & 78.47 (0.9) & 79.33 (0.4)  \\
     \cmidrule(r){2-8}
    &KL-Norm&\textbf{73.20} (0.5)  & \textbf{76.82} (0.8)  & \textbf{76.97} (0.3) & \textbf{79.12} (0.8) & \textbf{80.34} (0.7) \\ 
    &$\bm{\Delta}$ & +2.08 &  +1.52 &  +1.07 &  +0.94 & +0.83\\
    
    \toprule 
    \multirow{6}{*}{YELP} &BERT$_\text{Base}$ & \textbf{41.58} (0.3) & \textbf{44.02} (0.5)  & 45.54 (0.5) & 47.62 (0.9) & 47.92 (0.7) \\ 
     &~~+BN & 38.08 (0.3)  & 43.18 (0.7)& 44.56 (0.3) & 45.58 (0.3) & 47.04 (0.8)   \\
     &~~+LN & 41.13 (0.4)  & 43.37 (0.4)&46.01 (0.6) & 46.35 (0.5) & 47.94 (0.8)   \\
     &~~+GN & 40.34 (0.7) & 43.11 (0.6)  & 45.12 (0.2) & 45.79 (0.4) & 47.17 (0.3)  \\
    &~~+Dropout & 40.86 (0.3) & 43.37 (0.4) & 45.02 (0.9) & 46.77 (0.7) & 48.02 (0.2)  \\
    &~~+WD & 40.77 (0.9) & 43.60 (1.1) & 45.07 (0.6) & 47.12 (1.1) & 47.83 (0.9) \\ 
    \cmidrule(r){2-8} 
    &KL-Norm & 41.48 (0.4)  & 43.86 (1.1) & \textbf{46.10} (0.7) & \textbf{48.38} (0.4) & 
  \textbf{48.58} (0.4) \\
   &$\bm{\Delta}$ &-0.1 &  -0.16 &  +0.56 & +0.76 &  +0.66 \\
   \bottomrule
    \end{tabular}}
    \label{tab:sampled_results_test} \vspace{-0.5em} 
    \end{table*}

 \begin{table*}[t] 
    \caption{Test accuracies in the low-resource setting on speech classification under varying sizes of training data (300, 600, 900, 1200 and 1500 samples). $\bm{\Delta}$ shows the absolute difference between the results of the KL-Norm model with Wav2Vec-base.} 
    \centering
    \resizebox{12cm}{\height}{
    \vspace{-1em}
    \begin{tabular}{l|l|l|l|l|l|l|l}
    \toprule 
\textbf{Data}     &  \textbf{Model} & $\bm{300}$ & $\bm{600}$& $\bm{900}$ & $\bm{1200}$ & $\bm{1500}$ \\
    \toprule      
    \multirow{6}{*}{GoogleC} & Wav2vec & 63.21 (0.9)  & 83.15 (0.5)& 84.32 (0.7) & 85.58 (0.5) & 86.92 (0.4) \\ 
    &~~+BN & 64.70 (0.6)  & 84.99 (0.7)&86.81 (0.6) & 87.33 (0.7) & 88.17 (0.9)  \\
    &~~+LN & 63.86 (0.4)  & 84.11 (0.5)&85.68 (0.6) & 86.99 (0.2) & 87.73 (0.6)  \\
    &~~+GN & 63.14 (0.2) & 84.03 (0.2)  & 85.16 (0.4) & 86.45 (0.6) & 87.01(0.1)  \\
    &~~+Dropout & 63.11 (0.3)  & 83.45 (0.7)& 84.87 (0.4) & 85.63 (0.7) & 86.61 (0.4)  \\
    &~~+WD & 63.21 (0.8) & 83.56(0.6) & 84.93(0.5) & 85.77 (0.8) & \textbf{86.79} (0.4) \\ 
    \cmidrule(r){2-8}
    &~~+KL-Norm &\textbf{70.16} (0.9) &\textbf{87.17} (0.7) & \textbf{87.93} (0.8) & \textbf{88.44} (0.4) & \textbf{89.08} (0.3)\\
    & $\bm{\Delta}$ &  +6.95 &  +4.02 &  +3.61 &  +2.86 & +2.16& \\
    \toprule 
    \multirow{6}{*}{ESD} &Wav2Vec & 55.23 (0.7) & 62.02 (0.8)  & 72.56 (0.3) & 75.01 (0.4) & 77.12 (0.5) \\ 
     &~~+BN & 56.82 (0.6)  & 62.93 (0.6)& 73.67 (0.4) &  75.43 (0.8) & 77.88 (0.7)   \\
     &~~+LN & 56.36 (0.3)  & 62.58 (0.7)& 73.11 (0.8) & 75.19 (0.6) & 77.16 (0.4)   \\
     &~~+GN & 55.08 (0.3) & 62.11 (0.2)  & 72.36 (0.8) & 74.79 (0.6) & 76.52 (0.4)  \\
    &~~+Dropout & 54.15 (0.6) & 61.66 (0.8) & 71.56(0.7) & 74.33 (0.2) & 76.21 (0.4)  \\
    &~~+WD & 54.34 (0.7) & 61.79 (0.6) & 71.62 (0.4) & 74.72 (0.9) & 76.19 (0.8) \\ 
    \cmidrule(r){2-8} 
    &KL-Norm & \textbf{59.62} (0.7)   & \textbf{65.86} (0.8) & \textbf{75.67} (0.6) & \textbf{77.13} (0.2) & \textbf{78.53} (0.7)\\
   &$\bm{\Delta}$ &+4.39 &  +3.84 &  +3.11 & +2.12 &  +1.41 \\
    \toprule 
        \multirow{6}{*}{Crema} & Wav2vec & 52.12 (0.2) & 57.13 (1.2)  & 59.74 (0.5) & 63.76 (0.5) & 65.67 (0.6) \\
    &~~+BN & 52.07 (0.2)  & 58.03 (0.9)&59.76 (0.7) & 64.62 (0.8) & 65.26 (0.2)  \\
    &~~+LN & 51.38 (0.9) & 57.29 (0.8)  & 59.16 (0.7) & 64.01 (0.6) & 64.94 (0.7)  \\
    &~~+GN & 51.05 (0.9) & 57.12 (0.4)  & 59.21 (0.2) & 63.42 (0.8) & 63.79 (0.6)  \\
     &~~+Dropout & 50.61 (0.7) & 56.88 (0.8)  & 57.03 (0.8) & 62.67 (0.6) & 63.08 (0.7)   \\ 
    &~~+WD& 50.88 (0.7) & 56.91 (0.7)  & 57.28 (0.6) & 62.84 (0.2) & 63.19 (0.4) \\
     \cmidrule(r){2-8}
     &~~+KL-Norm & \textbf{55.45} (0.8)  & \textbf{58.95} (1.3) & \textbf{61.21} (0.9) & \textbf{64.89} (0.8) & \textbf{66.57} (0.6) \\
     & $\bm{\Delta}$ & +3.33 & +1.82 & +1.47 & +1.13 & +0.90\\ 
   \bottomrule
    \end{tabular}
    }
   
    \label{tab:sampled_results_test_speech} \vspace{-0.5em} 
    \end{table*}

\subsection{Analysis of Out of domain Generalization}\label{sec:genera}

We have used various NLP datasets to see the effectiveness of KL Regularized normalization in the out of domain generalization .We have used datasets referred in~\cite{karimi2020bias}, including SICK~\cite{MARELLI14.363}, ADD1~\cite{pavlick2016most}, JOCI~\cite{zhang2017ordinal}, MPE~\cite{lai2017natural},  MNLI, SNLI, SciTail~\cite{khot2018scitail}, and three datasets from \cite{white2017inference} namely DPR~\cite{rahman2012resolving}, FN+~\cite{pavlick2015framenet+}, SPR~\cite{reisinger2015semantic}, and Quora Question Pairs (QQP) interpreted as an NLI task as by ~\cite{gong2017natural}.  We use the same split used in~\cite{wang2017bilateral} for the experiment. We have trained the model on 6000 samples of SNLI and MNLI datasets and used these dataset as the test set to see the out of domain generalization of the proposed model.The SNLI and MNLI datasets contain three labels of contradiction, neutral, and entailment. However, some of the considered target datasets have only two labels, such as DPR or SciTail. When the target dataset has two labels of \emph{entailed} and \emph{not-entailed}, as in DPR, we consider the predicted contradiction and neutral labels as the not-entailed label.  In the case the target dataset has two labels of \emph{entailment} and \emph{neutral}, as in SciTail, we consider the predicted contradiction label as neutral.

Table \ref{tab:transfer_results} shows that KL-Norm gives an average improvement of 2.62\% and 4.5\% over accuracy when trained with SNLI and MNLI respectively from BERT$_\text{Base}$ model. It has shown substantial improvement against all other baseline models.  These results support our claim that KL-Norm motivates learning more general features, rather than redundant superficial features, leading to an improved generalization to datasets without these superficial biases.

\begin{table*}[t!] 

 \caption{Test accuracy of models transferring to new target datasets. All models are trained on SNLI or MNLI and tested on the target datasets. \bm{$\Delta$} are absolute differences with BERT$_\text{Base}$.}
    \centering
    \resizebox{1.0\textwidth}{!}{
    \begin{tabular}{lllllllllllllll}
        \toprule 
        & \multicolumn{5}{c}{\textbf{SNLI}} & \multicolumn{5}{c}{\textbf{MNLI}} \\ 
         \cmidrule(r){2-7} \cmidrule(l){8-13} 
        {\vspace{-0.75em} \textbf{Data}} &  \multicolumn{2}{c}{} & \multicolumn{2}{c}{}\\ 
                 & \rotatebox{0}{\textbf{\footnotesize{BERT$_\text{Base}$}}} & \rotatebox{0}{\textbf{\footnotesize{~~+BN}}} & \rotatebox{0}{\textbf{\footnotesize{~~+LN}}}  &
                 \rotatebox{0}{\textbf{\footnotesize{~~+GN}}} &
                 \rotatebox{0}{\textbf{\footnotesize{+KL-Norm}}}& \bm{$\Delta$} &\textbf{\footnotesize{BERT$_\text{Base}$}} & \textbf{\footnotesize{~~+BN}} & \textbf{\footnotesize{~~+LN}}&
                 \textbf{\footnotesize{~~+GN}}&
                 \textbf{\footnotesize{+KL-Norm}}& \bm{$\Delta$}\\  
        \toprule  
        JOCI & 46.03 & 47.13& 44.67 & 45.55& 51.82 & +5.79 & 46.41 & 48.12 &45.21 & 44.31 & 54.11 & +7.7\\
        ADD1 & 45.61& 38.75 &44.7  & 39.81 &44.14 & -1.47 & 51.22 &35.65 & 46.42 & 47.21 & 56.67 &+5.45 \\
        DPR & 49.22 & 49.12&49.18 &49.11  &50.31 & +1.09 & 49.95 &49.31 & 49.4 & 49.3 &50.11 & +0.16\\
        SPR& 37.07  & 35.43 & 37.48 &35.55 &36.12 &+1.7 & 42.37  & 40.45 & 45.12 &42.11 &43.17& +0.80 \\
        FN+ & 45.31 & 50.61 & 44.31 & 47.71 &47.35  &-0.95 & 43.5 &43.2 &43.53 & 43.3&44.88 & +1.38 \\
        SICK & 53.06 &46.78 &53.98 & 45.11 &55.61  &+2.55 & 65.07 &62.43 &66.60 &63.33 & 71.64 & +6.53 \\
        MPE &  58.12 & 57.21 &57.44  & 54.33 &63.61  &+5.48 & 55.10 &54.43 & 57.5 & 56.48& 58.31 & +3.21 \\
        SCITAIL & 64.81 &58.13 &65.23 & 58.43 & 70.88  &+6.07 &64.67 &67.49 & 66.61 &64.31 &75.43 & +10.76  \\
        QQP & 62.91 & 60.93& 59.88  & 58.07 & 65.87  &+2.96 &63.38 &60.16 &62.89 & 60.11 & 69.67& +6.29 \\
        SNLI Hard & 64.39&63.12 &65.37  &63.31  &67.45  &+3.06 &54.11 &53.78 & 54.38 &53.55 & 56.91 & +2.8\\
        \midrule 
        Average &--- & ---& --- & --- & --- & +2.62 & --- & --- & --- & --- & --- & +4.50 \\ 
    \bottomrule 
    \end{tabular}}
    \label{tab:transfer_results}
\end{table*}

 \subsection{Impact of KL-Norm on Overfitting}\label{sec:overfit-vi}
\begin{figure}[h!]
\centering
  \includegraphics[width=\linewidth]{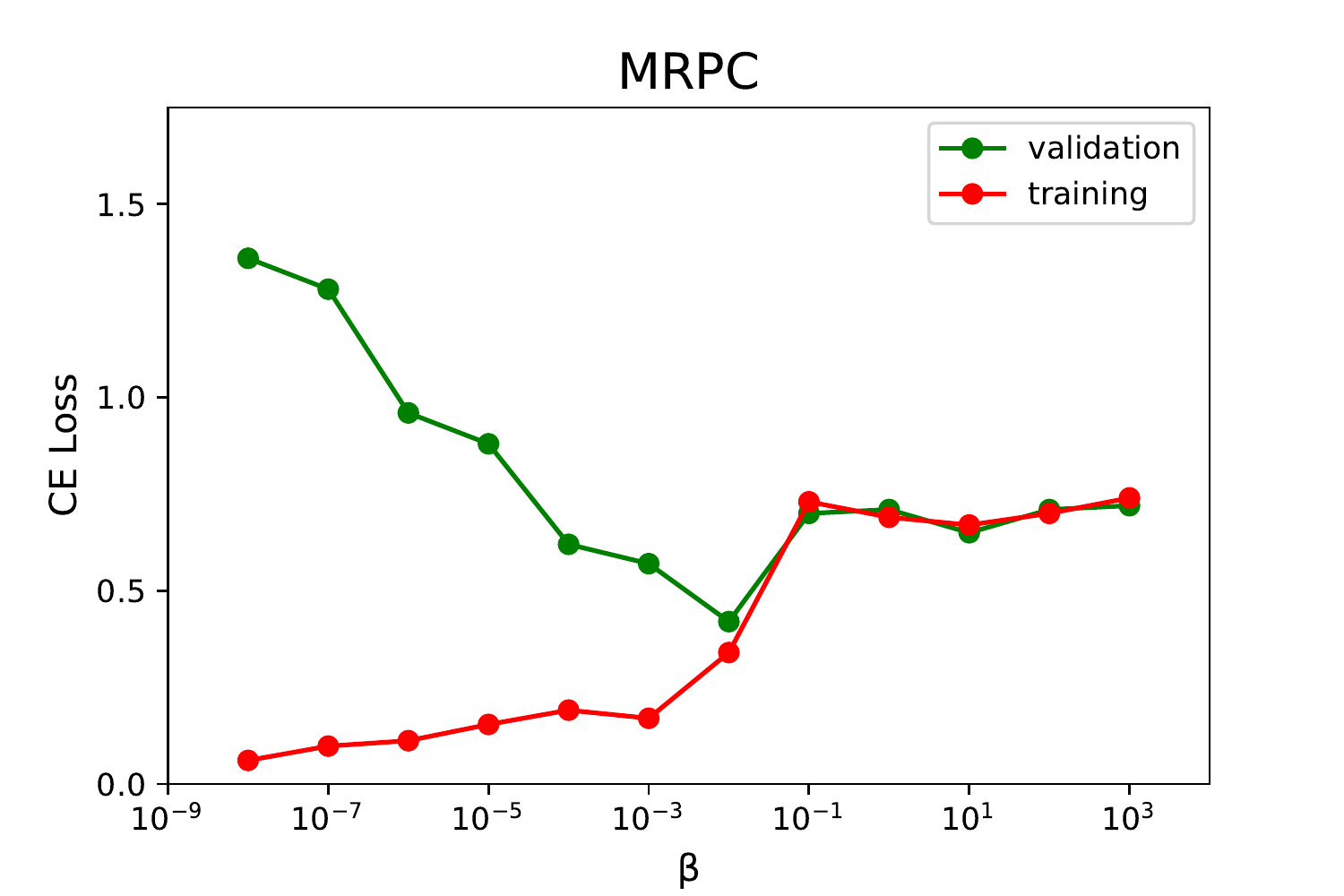}
  \caption{ Validation and training losses of KL-Norm for varying $\beta$ and a fixed bottleneck size on GLUE.}
  \label{fig:mrpc}
\end{figure}

\begin{figure}[h!]
\centering
  \includegraphics[width=\linewidth]{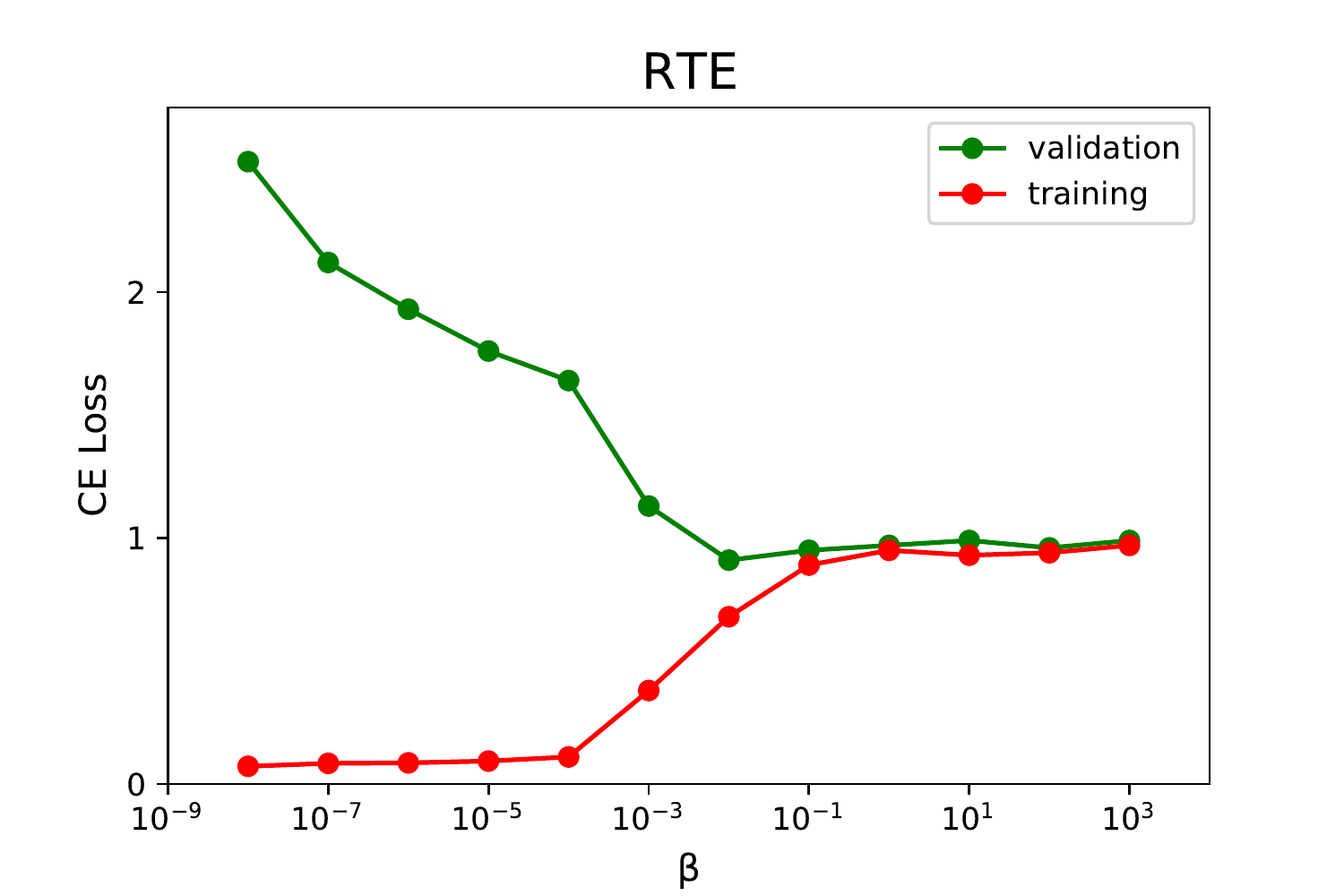}
  \caption{ Validation and training losses of KL-Norm for varying $\beta$ and a fixed bottleneck size on GLUE.}
  \label{fig:rte}
\end{figure}

 Loglikelihood and KL-divergence are typically balanced by a suitable $\beta$-parameter, since they have somewhat contrasting effects: the former will
try to improve the quality of the reconstruction, neglecting the shape of the latent space; on the other side, KL-divergence is normalizing and smoothing the latent space. Tuning the $\beta$ parameter is crucial to reduce the over-fitting.
We analyze the effect of KL-Norm on the generalization of model and in reducing overfitting. We analyze the effect of the $\beta$ parameter on training and validation error. We fix the bottleneck size ($K$) based on the models selected in Section~\ref{sec:glue_results}, and we train KL-Norm  model on the GLUE benchmark for varying values of $\beta$ and plot the validation and training loss in Figure~\ref{fig:mrpc} and \ref{fig:rte}. 

KL-Norm has little effect for the small value of $\beta$ as the validation loss is substantially higher than training loss showing the case of over-fitting. This is because network become too deterministic ($\Sigma \approx 0$) and learns irrelevant features not needed to predict the labels.As we increase the $\beta$ , we see the better generalization of the network. As the $\beta$ become too large, again the validation loss increases as it starts blocks relevant features needed to predict the labels.

\subsection{Analysis of Removal of irrelevant features}\label{sec:iree}
We have used \cite{elazar2018adversarial,Belinkov2019OnAR} framework to evaluate whether the debiasing methods are succesful in removing the biases from sentence or not.
 After debiasing, the trained encoder is frozen and the classifier is retrained to try to extract the biases.  If the classifier reaches high accuracy given only bias features, then the encoder's representation has not been successfully debiased.
 We train a classifier which only sees the representation of the hypothesis sentence and see if it can predict the class of the sentence pair, which is an established criterion to measure known biases in NLI datasets~\cite{gururangan2018annotation}. Thus, we freeze the trained encoders from our model and the BERT baseline and retrain a hypothesis-only classifier on hypotheses from the SNLI and MNLI datasets.For reference, we compare to a hypothesis-only model with a BERT encoder trained end-to-end.  Table~\ref{tab:bias_results} shows the results which shows that KL-Norm is able to achieve lower accuracy against all baseline models. 

 \begin{table}[t!]
    \caption{Hypothesis-only accuracy when freezing the encoder from models trained on SNLI/MNLI in Table~\ref{tab:sampled_results_test} and retraining a hypothesis-only classifier, and baseline results when the encoder is not frozen (H-only). Lower results show more successful debiasing.
    \vspace{-0.5ex}}
    \centering
    \begin{tabular}{lllllll}
    \toprule 
     \multirow{2}{*}{\textbf{Model}} & \multicolumn{3}{c}{\textbf{SNLI}} & \multicolumn{3}{c}{\textbf{MNLI}} \\ 
         \cmidrule(r){2-4} \cmidrule(l){5-7} 
      &   \textbf{Train}  & \textbf{Dev} & \textbf{Test} &  \textbf{Train}  & \textbf{Dev} & \textbf{Test} \\
     \toprule
    H-only & 95.2 & 56.62 & 55.78 & 83.12 & 51.34 & 51.12\\ 
    \midrule 
    BERT$_\text{Base}$ &   73.2 & 51.87 &  51.17  &  52.7 &  42.68 & 43.55\\
    ~+BN &   71.1 & 51.91 &  51.11  &  58.5 &  44.68 & 44.03\\
    ~+LN &   70.9 & 51.98 &  52.26  &  58.5 &  44.68 & 44.03\\
    ~+GN &   71.9 & 52.75 &  53.08  &  58.7 &  44.91 & 44.72\\
    ~+KL-Norm &  49.5 &  \textbf{41.41} &   \textbf{40.12} & 37.4 &  \textbf{35.28} & \textbf{35.99} \\
    \bottomrule
    \end{tabular}

    \label{tab:bias_results}
\end{table} 

\subsection{Analysis of model parameters}\label{sec:mempara}
Table \ref{tab:performance} shows the efficiency evaluation of the proposed model in terms of number of model parameters and  memory overheads with K = $512$. The peak memory overheads increases by 1.92\% against all other baseline model. KL-norm has substantially lower memory overheads as compared to weight decay. VI-Norm is useful while dealing with large scale transformer model such BERT, ROBERTA, etc. There is a negligible increment(1.68\%) of the model parameters due to addition of MLP layers to calculate the affine parameters i.e. shift and scale. 

\begin{table}
    \caption{Performance evaluation for all methods. $\bm{\Delta\%}$ are relative differences with BERT$_\text{Base}$.}\vspace{-0.5em}
\resizebox{8cm}{\height}{
    \begin{tabular}{l|lr|lr}
    \toprule 
    \textbf{Model} & \textbf{Memory} & \bm{$\Delta$\%} & \#\textbf{Parameters} &\bm{$\Delta$\%} \\
    \toprule 
    BERT$_\text{Base}$      & 418.74 GB  &---       & 109.48 M & ---  \\
    ~+BN   & 418.74 GB  &0\%        & 109.48 M & 0\%   \\
    ~+LN  & 418.74 GB  &0\%       & 109.48 M & 0\%   \\
    ~+GN  & 418.74 GB  &0\%       & 109.48 M & 0\%   \\
     ~+WD      & 506.19 GB  & 20.88\%  & 109.48 M & 0\% \\ 
     ~+Dropout & 418.74 GB  & 0\%      & 109.48 M & 0\%  \\ 
     \midrule 
     ~+KL-Norm & 426.80 GB  & 1.92 \% & 111.33 M & 1.68\% \\
    \bottomrule
    \end{tabular}}

\label{tab:performance}
\end{table}

\subsection{Ablation Study}
\paragraph{Analysis of model without KL loss}
 Table~\ref{tab:glue_results_ablation} shows the evaluation on three GLUE datasets  without regularization loss($\beta$ =$0$). The architecture just reduces to deterministic dimensionality reduction with an MLP. This evaluation shows the performance increment while adding the regularization loss.
 
\begin{table}[t!]
 \caption{Average ablation results over 5 runs with std in parentheses on GLUE.} 
\centering
\resizebox{6cm}{\height}{
    \begin{tabular}{l|l|l|l|l|l|l|l|l} 
        \toprule 
         & \multicolumn{2}{c}{\textbf{MRPC}} & \multicolumn{1}{c}{\textbf{RTE}}\\ 
         \cmidrule(r){2-3} \cmidrule(l){4-4}
        {\vspace{-0.75em} \textbf{Model}} &  \multicolumn{2}{c}{} & \multicolumn{2}{c}{} &\\ 
                 & \textbf{Accuracy} & \textbf{F1} & \textbf{Accuracy}\\  
        \toprule 
    BERT$_\text{Base}$ & 84.31 (0.2) & 89.01 (0.2)  & 64.23 (1.8) \\ 
    ~~+BN & 86.76 (0.5) & 90.49 (0.5)  & 66.42 (1.4) \\
    ~~+LN & 85.23 (0.3) & 88.12 (0.7)  & 65.17 (0.9) \\
    ~~+GN & 85.01 (0.2) & 87.98 (0.5)  & 65.55 (0.2) \\
    \midrule 
    ~~+KL-Norm ($\beta$=0) &86.27 (0.4) &90.03 (0.3)  & 67.17 (0.8) \\ 
   ~~+KL-Norm & \textbf{87.25 (0.1)} & \textbf{91.03 (0.6)} & \textbf{70.42 (0.8)} \\ 
    \bottomrule
    \end{tabular}}
    \label{tab:glue_results_ablation} 
\end{table}

\paragraph{Analysis of model in high resource setting}
We have done the experiments win high resource setting with two NLP datasets (MNLI and QNLI)and one speech datasets(Google speech Command) to see the behaviour of the proposed model . We found the results are comparable with other normalization techniques as shown in table \ref{tab:table45}. The reason is the learnable parameters of traditional normalization techniques is able to make the data well behaved with the proposed KL regularized normalization in higher resource settings.

\begin{table}[h!]
\caption{Average accuracy over 5 runs with std in parentheses in High resource setting}
    \label{tab:table45}
  \begin{center}

    \begin{tabular}{c|c|c|c} 
      \textbf{Model} & \textbf{MNLI} & \textbf{QNLI} & \textbf{GoogleC}\\
      \hline
       BERT$_\text{Base}$ & 82.92(0.24) & 90.12(0.12) & - \\
        ~~+BN & 82.23 (0.21)  & 89.14(0.23)  & - \\
        ~~+LN & 82.98 (0.18) & 90.34 (0.14)  & -\\
        ~~+GN & 82.01 (0.13) & 90.01 (0.15)  & - \\
        ~~+KL-Norm  &82.27 (0.28) &90.03 (0.09)  & - \\
        \midrule 
        Wav2Vec & - & - &  94.78(0.12) \\
        ~~+BN & - & - & 93.12(0.13) \\
        ~~+LN & - & - & 94.84(0.17) \\
        ~~+GN & - & - & 94.52(0.31) \\
         ~~+KL-Norm & - & - & 94.29(0.21) \\

    \end{tabular}
 
  \end{center}
\end{table}

\section{Conclusion}
In this paper, we have proposed a novel KL Regularized normalization framework, KL-Norm, that calculates the rescaling parameters of normalization along with imposing the gaussian prior through KL loss. It incorporates stochasticity which gives ensemble effect that helps the model to give higher accuracy. Addition of KL loss acts as a regularizer that reduces overfitting and better generalization. It removes the irrelevant features of data. This approach is based on density estimation which performs better on out of domain generalization.
Experimental evaluations on low resource downstream NLP tasks using pre-trained BERT and speech task using pre-trained Wav2Vec2.0 model demonstrate superior performance of the proposed framework against baseline models.

\end{document}